\documentclass[10pt,twocolumn,letterpaper]{article}

\usepackage{cvpr}
\usepackage{times}
\usepackage{epsfig}
\usepackage{graphicx}
\usepackage{amsmath}
\usepackage{amssymb}

\usepackage{multirow}
\usepackage{rotating}

\usepackage{algorithm}%
\usepackage{algpseudocode}%

\usepackage[pagebackref=true,breaklinks=true,letterpaper=true,colorlinks,bookmarks=false]{hyperref}

\usepackage[colorinlistoftodos]{todonotes}

\cvprfinalcopy %

\def\cvprPaperID{17} %

\ifcvprfinal\fi
\begin{document}

\title{Speeding Up Neural Networks for Large Scale Classification using WTA Hashing}

\author{Amir H. Bakhtiary\\
Universitat Oberta de Catalunya\\
Email: abakhtiary@uoc.edu
\and
Agata Lapedriza\\
Universitat Oberta de Catalunya\\
Email: alapedriza@uoc.edu
\and
David Masip\\
Universitat Oberta de Catalunya\\
Email: dmasipr@uoc.edu
}

\maketitle
\begin{abstract}

In this paper we propose to use the Winner Takes All hashing technique to speed up forward propagation and backward propagation in fully connected layers in convolutional neural networks. The proposed technique reduces significantly the computational complexity, which in turn, allows us to train layers with a large number of kernels with out the associated time penalty.

As a consequence we are able to train convolutional neural network on a very large number of output classes with only a small increase in the computational cost. To show the effectiveness of the technique we train a new output layer on a pretrained network using both the regular multiplicative approach and our proposed hashing methodology. Our results showed no drop in performance and demonstrate, with our implementation, a 7 fold speed up during the training. 

\end{abstract}

\section{Introduction}

Convolutional Neural Networks, CNN, have recently achieved state of the art performance in a number of computer vision tasks \cite{krizhevsky_imagenet_2012,razavian2014cnn}.  Even more remarkably, it has been shown that using the output of 7th layer of the network (FC7) when trained on the ImageNet benchmark \cite{deng_new_2013}  as a generic feature descriptor it is possible to achieve higher performance than traditional handcrafted features from the computer vision literature \cite{razavian2014cnn}. 

One of the main challenges when working with large networks is the  computational cost during training. In fact, many different efforts have emerged to increase the speed of these networks. This includes using special commands on CPUs~\cite{vanhoucke_improving_2011}, using GPUs (as in frameworks such as Torch, Caffe and Theano), and using FPGAs~\cite{farabet_large-scale_2011}~\cite{nagy_configurable_2003}. Although these techniques have had great success in reducing training and testing time for deep neural nets, the amount of computations required has remained constant. This is because they have focused on executing these operations faster and/or in parallel, instead of reducing the number of needed computations.

\begin{figure}[!t]
\centering
\includegraphics[width=3.2in]{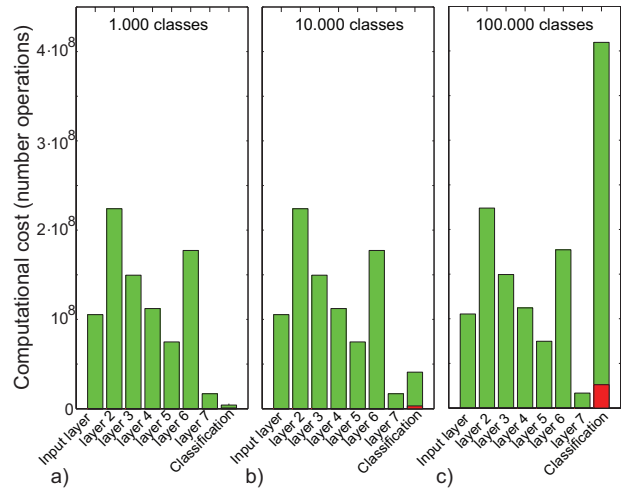}
\caption{Distribution of computational cost on AlexNet \cite{krizhevsky_imagenet_2012} as we vary the number of output classes from (a) 1,000, (b) 10,000 and (c) 100,000. In red we show the computational cost with our proposed method.}
\label{teaser}
\end{figure}

Figure \ref{teaser}.a shows the distribution of the computational cost of the deep neural network from \cite{krizhevsky_imagenet_2012} which is the basis of many of current approaches in computer vision. Although the computational cost of the classification layer in these approaches is small compared to the rest of the network, this can change easily as the number of classes increases. The reason for this is that the number of  kernels in a classification layer scales linearly with the number of classes.

For example, as we increase the number of classes  in  (figure.~\ref{teaser}.a) to 10,000 (figure~\ref{teaser}.b) or 100,000 (figure~\ref{teaser}.c) the cost of the output layer quickly dominates the computational cost of the network. Furthermore, this cost would become even more pronounced if the output size of layer 7 is increased, since, the computation complexity of a layer is linear in both the input size and the output size.

In this paper we introduce a hashing-based technique that dramatically reduces the computational cost of training and testing of sparse fully connected convolutional layers. The red bar in figure.~\ref{teaser}.b-c shows the computational cost of computing the output when using our approach. This improvement allows using the same network architecture on a very large number of classes without significantly increasing the computational cost.

To this end we make use of $2$ main observations: (1) the most computationally expensive operation for these fully connected layers is matrix multiplication which occurs both during forward propagation and backpropagation, and (2) when propagating through layers whose output is trained for classification, units that have low activation would have  performed computations that will not be used. This is because we are only interested in units that are "turned on" and which will lead to changes in the output. The same is true for backpropagation, units with extremely low values will have very low slope and can be ignored or aggregated. This reasoning holds true for any layer with a non-linearity function that applies a cut off on the output values but the extra computation is more pronounced in layers that are trained to be sparse. 

Although several methods have been proposed for more efficient approaches to matrix multiplication~\cite{li_strassens_2011} these have not been yet applied to neural networks and, in most cases, current implementations only lead to a modest saving in computational complexity of $O(n^{2.807})$ vs $O(n^3)$. As a consequence, the current implementations of neural networks have a computational cost that grows linearly with the number of units. One place where this cost is very tangible is the final classification layer where the number of units corresponds to the number of expected classes. 

Here we propose to use winner take all (WTA) hashing to identify the units that will have sufficiently high amplitude before performing any expensive computation. Then, only for the identified units, we will compute the exact output of these units, while using a default low value for the rest of the units. This way only a small number of elements in the layer output matrix need to be computed while the rest are set to a predefined value. This WTA technique has been already used in vision tasks. In particular,it was successfully applied on a HOG-based object detector \cite{dean2013fast} in order to detect 100.000 object classes.

Furthermore, during the backpropagation phase, we only backpropagate through units that were activated. This is similar to using drop out. We also backpropagate through units that were not activated but should have been. 

The reason for backpropagating through activated units is to minimize the computational cost. The expected output, after the non-linearity layer, is sparse.  Also, the gradient of the loss will have the same sparsity pattern. Therefore, most of the errors fed back on the non-activated nodes would actually be near zero or exactly zero, depending on the type of non-linearity used. 

To underline the significance of this approach, the application of this technique for training neural networks to classify into 10k classes is shown and a 7 fold speed-up in training the layer is demonstrated.

\section{Winner takes all hashing}

Winner Take All (WTA) hashing is a method that transforms the input feature space into binary codes. In the resulting space Hamming distance closely correlates with rank similarity measure \cite{yagnik2011power}. The obtained binary descriptors show a degree of invariance in front of slight perturbations of the original data, which makes the method a suitable basis for retrieval algorithms. The WTA is based on random permutations on the data components and does not require any data-driven learning.

\begin{figure}[!t]
\centering
\includegraphics[width=2.5in]{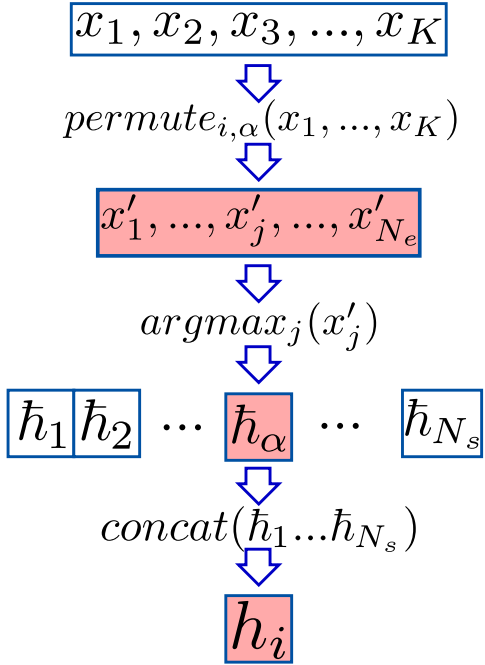}
\caption{Computation of one hash for one vector.  We first permute the vector that is to be hashed. Then we compare the first $N_e$ elements. The index of the highest element is recorded. concatenation of $N_s$ indexes derives one hash value.}
\label{fig_hash}
\end{figure}

 Figure  \ref{fig_hash} summarizes the computation of one WTA hash. First we permute the input vector $[x_1,x_2,\dots,x_K]$ by indexing the incoming data through the permutation arrays. A different permutation is used for each section $\alpha$ and for each hash $i$. In each section the $N_e$ first elements of the permuted vector is selected:$[x'_1,x'_2,\dots,x'_{N_e}]$. These are compared and the index of the largest element is recorded, $\hbar_\alpha$. We compute $N_s$ sections for each hash. These sections are all concatenated to form one single hash $h_i$.

For instance, if we have a vector ${\bf x}=[0.2,0.9,0.4,0.5,0.1]$, and use a two-band hash with permutations $P_1=[3,2,4,5,1]$ and $P_2=[4,1,5,3,4]$, and $n_b=4$, the first indices of each permutation are $[3,2,4,5]$, $[4,1,5,3]$ respectively. The selected elements are $[0.4,0.9,0.5,0.1]$ and $[0.5,0.2,0.1,0.4]$. The maximum values are $0.9$ and $0.5$, and the respective indices $2$ and $1$. The resulting concatenated band is [0110] (least significant bit leftmost).

\section{Speeding up Neural Networks computations with WTA hashing}

Neural networks layers usually have sparse activation patterns. The output values of the units in these layers are mostly a constant value, usually zero, while a few units will have output values that are larger. These patterns are obtained either explicitly using sparse regularization terms or implicitly because of the data that the layer is trying to learn.

L1 regularization is an example of training that explicitly leads to sparsity. In contrast, an example of sparsity resulting from data includes training the classification layer, where the expected data is sparse. 

When training a layer to be sparse, the non-linear function following the multiplication has no parameters to learn and the sparsity constraint is fed back to the weights involved in the multiplication. How the weights achieve sparsity depends on the type of non-linearity layer used. 

Usually the non-linear function has an output of zero or near zero for values that are less than a given threshold. To achieve sparsity in these layers, the weights would learn to arrive at outputs that are mostly less than the threshold value and only occasionally higher than the threshold. Therefore, to be able to correctly compute the output of the non-linear layer, we hypothesize that only the elements that are larger than the threshold need to be identified and computed.

The same concept can be applied to the backpropagation of the error term. As the gradient of the non-linearity is extremely small and possibly zero before the threshold, only the error values coming from the activated units need to be computed. In this paper we propose an efficient learning algorithm that ignores the computations related to the backpropagation of the remaining units.

Correspondingly, the most computationally expensive part of activation propagation in a layer of units is the multiplication of weights of the layer by the inputs of that layer and summing it together. Conversely, the most computationally expensive operations during back propagation is multiplication of the output error by the weights to obtain the input error and the multiplication of the output error by the input values to obtain the weight errors. The next subsections will describe how our approach speeds up these two computationally expensive parts of forward and backpropagation by identifying units that will have large values and only computing the values and errors of these units.

Note that in this work we are concerned with propagating batches of inputs through the layers. The reasons for this are the following:

\begin{enumerate}

\item The training algorithm used for the neural network is the stochastic gradient mini-batch descent, which is quite popular in the current literature. In this approach, at each iteration, the gradient of the loss of the network is computed over a batch of the training samples and a step is taken in the opposite direction of the gradient.

\item Processing a batch of inputs favors the computational economy. Although the runtime order of complexity remains the same there can be significant constant gains depending on the architecture used for the computation. This is more pronounced for parallel architectures such as GPUs.

\item The hashing approach is most efficient when batches of inputs are processed since, as with other hash based algorithms, a rehash is required every time the weights in the layer change. This will be elaborated on when discussing complexity in section \ref{section_complexity}.

\end{enumerate}

\subsection{Forward propagation}

\begin{figure*}[ht!]
\centering
\includegraphics[width=6in]{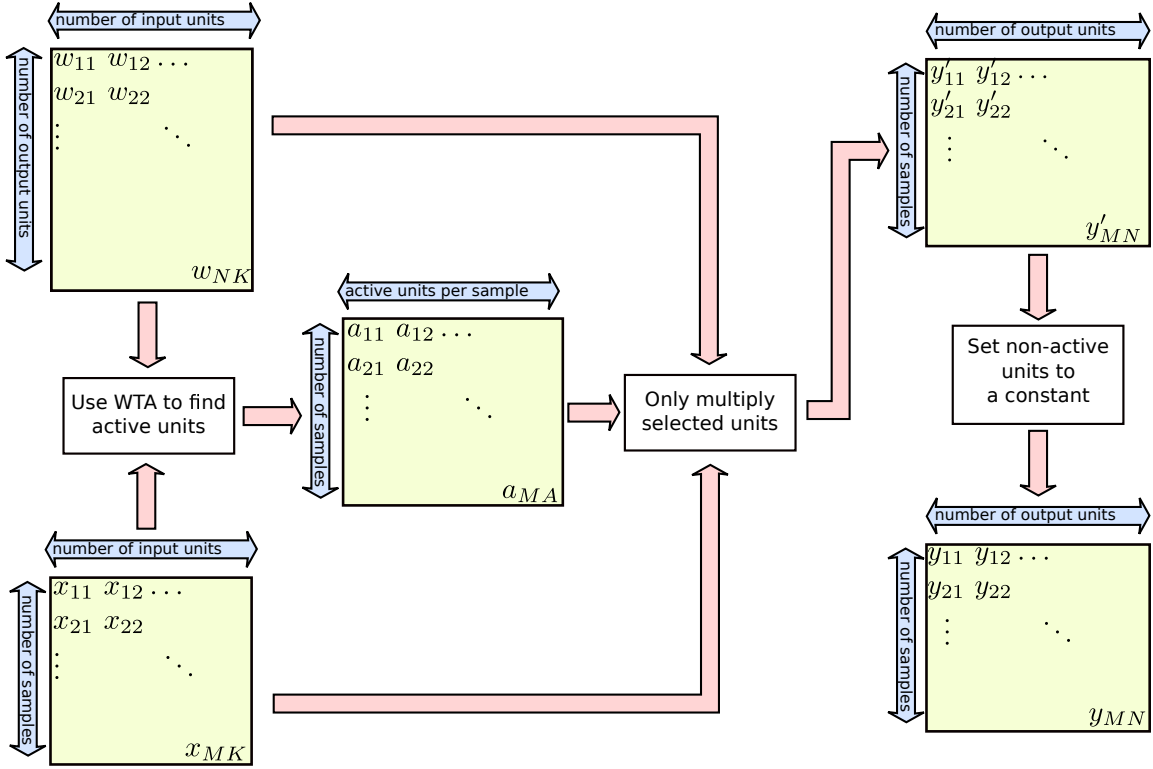}
\caption{Forward propagataion: calculating the output of a layer using the WTA hashing.}
\label{fig_hashedForward}
\end{figure*}

In a normal neural layer, propagation of one batch of $M$ samples through a layer with $N$ output units requires the computation of the value of the output units $y_{ji}$ for each pair of output unit $i\in\{1,...,N\}$ and sample $j\in{\{1,...,M\}}$. This is done by summing the product of the $K$ weights of the $j$th unit by the $K$ input unit values for the $i$th sample as is demonstrated in
\begin{equation}
\label{eq_summation}
y_{ji} = \sum_{k\in\{1,...,K\}} w_{ik}x_{jk}, 
\end{equation}
where $w_{ik}$ is the network parameter corresponding with the link connecting the $k$th input unit with the $i$th output, and $x_{jk}$ is the input value of the $k$th input unit for the $j$th sample.

The formulation given in equation~\ref{eq_summation} for computing the forward pass can be succinctly expressed in a matrix multiplication, one form of which can be
\begin{equation}
\label{eq_matrixMul}
\mathbf{Y = XW^\top},
\end{equation}
where $\mathbf{Y}$, $\mathbf{W}$, and $\mathbf{X}$ represent the aggregation of the outputs, the layer parameters and the inputs in matrix format.

In our approach, we propose to calculate only a little portion of the matrix multiplication in equation~\ref{eq_matrixMul}. We are interested in finding the elements in each row of the output that have the highest value. These elements correspond to columns in the $\mathbf{W^\top}$ matrix that result in a high value when multiplied a specific row in the $\mathbf{X}$ matrix. 

To do this we first use WTA hashing to identify units (columns in $\mathbf{W^\top}$) that will be active for each row of matrix $\mathbf{X}$. Figure~\ref{fig_hashedForward} depicts an overview of our proposed approach. We represent actived units by $a_{js}$ which stores the index of the $s$th activated unit for the $j$th sample. 

After the activated units are found, only the values corresponding to these units are computed for each sample. That is, for a given sample $j$, corresponding to a row in matrix $\mathbf{X}$, that row is only multiplied by the columns $a_{js}; s\in{1..S}$ in $\mathbf{W^\top}$. The effectiveness of the method is highlighted by contrasting this procedure against the full matrix multiplication where each row in $\mathbf{X}$ is multiplied by all columns in $\mathbf{W^\top}$.

\subsection{Selection of active units}

\begin{figure*}[!t]
\centering
\includegraphics[width=7in]{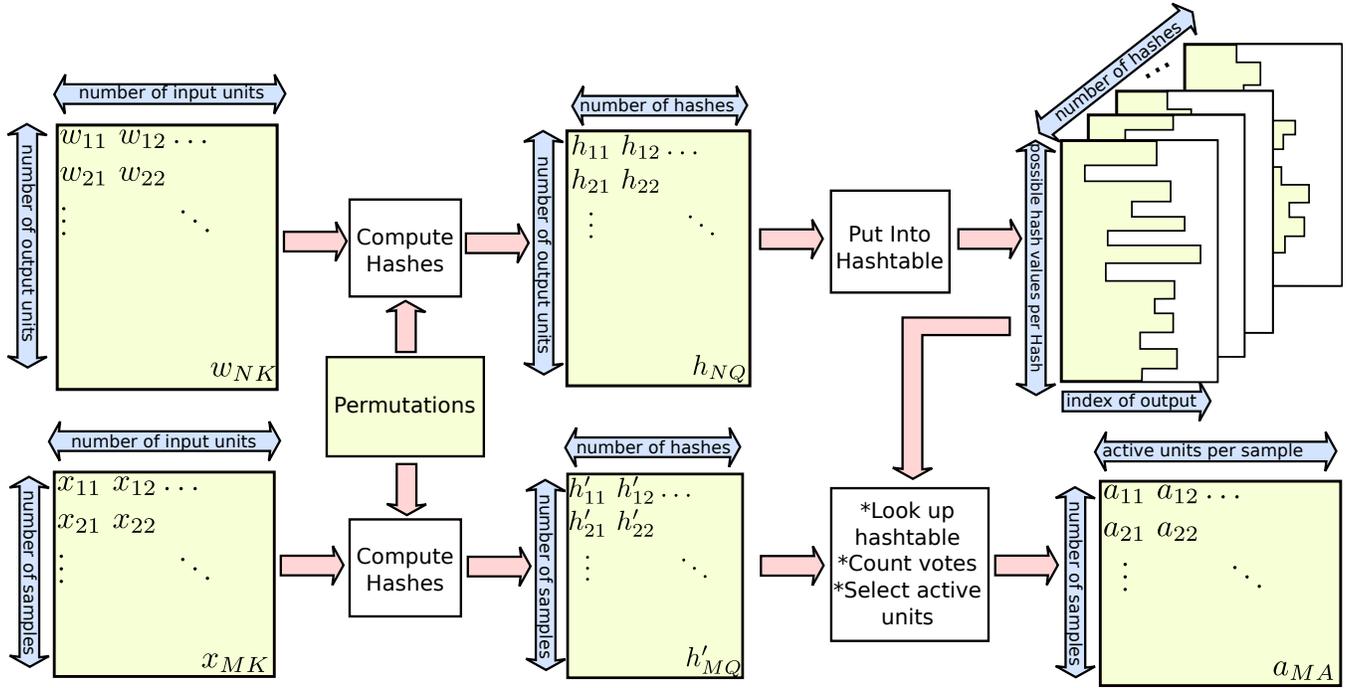}
\caption{Finding the activated unit.}
\label{fig_activeNeurons}
\end{figure*}

Figure \ref{fig_activeNeurons} outlines the process of computing the indexes of the active units $a_{js}$ for the $j$th sample. To do this first the WTA hashes of the units are computed. All the hashes are computed in a similar fashion, but according to different permutation arrays. Subsequently, we obtain $Q$ different hash values for each input vector. In the provided figure, $h_{iq}$ represents the hash value of the weights of the $i$th unit as given by the $q$th hash.

After computing the hash values for all the weights of the units, a multi hash table is constructed for each of the Q hashes. In each multi hash table $q$ the id of the $i$th unit is assigned to bin $h_{iq}$. Therefore each hash partitions the set of units according to their corresponding hash values under that permutation. This binning process is repeated for each permutation array so as to arrive at a set of Qhash tables.

After binning, the same permutation arrays are used to compute the hashes of the input samples leading to the values $\hbar_{jq}$. These hash values are then used to lookup the bins in their corresponding hash table. The contents of the retrieved bins constitute the votes for possible active units. For each input sample these votes are counted and the top voted units are selected as the active units. In the figure~\ref{fig_activeNeurons} the $a_{ju}$ is the index of the $u$th activated unit for the $j$th input sample. A total of $A$ units are chosen per input 

\subsection{Backward propagation}

\begin{figure*}[!t]
\centering
\includegraphics[width=7in]{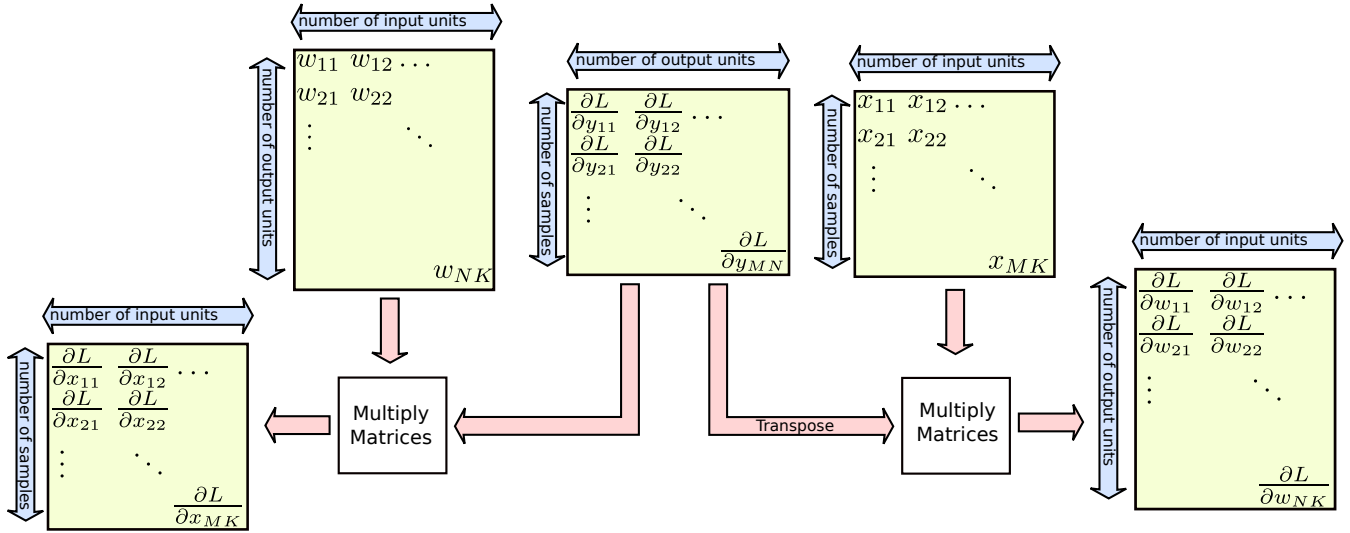}
\caption{Regular backpropagation}
\label{fig_normalBack}
\end{figure*}

When backpropagating the error terms, we are faced with the two matrix multiplications depicted in figure \ref{fig_normalBack}. One of this multiplications is performed to estimate the the partial derivatives of the loss with respect to the weights $\frac{\partial L}{\partial w_{ik}}$ for each output unit $i$ and each input unit $k$. The other multiplication estimates the partial derivatives of the loss with respect to the inputs $\frac{\partial{L}}{\partial(x_{jk})}$ for each input sample $j$ and each input unit $k$:
\begin{equation}
\frac{\partial L}{\partial w_{ik}} = \sum_j \frac{\partial L}{\partial y_{ji}}x_{jk}
\end{equation}
\begin{equation}
\frac{\partial L}{\partial x_{jk}} = \sum_i \frac{\partial L}{\partial y_{ji}}w_{ik},
\end{equation}
where $\frac{\partial L}{\partial y_{ji}}$ is the loss wrt the values of the output units.

\begin{figure*}[!t]
\centering
\includegraphics[width=7in]{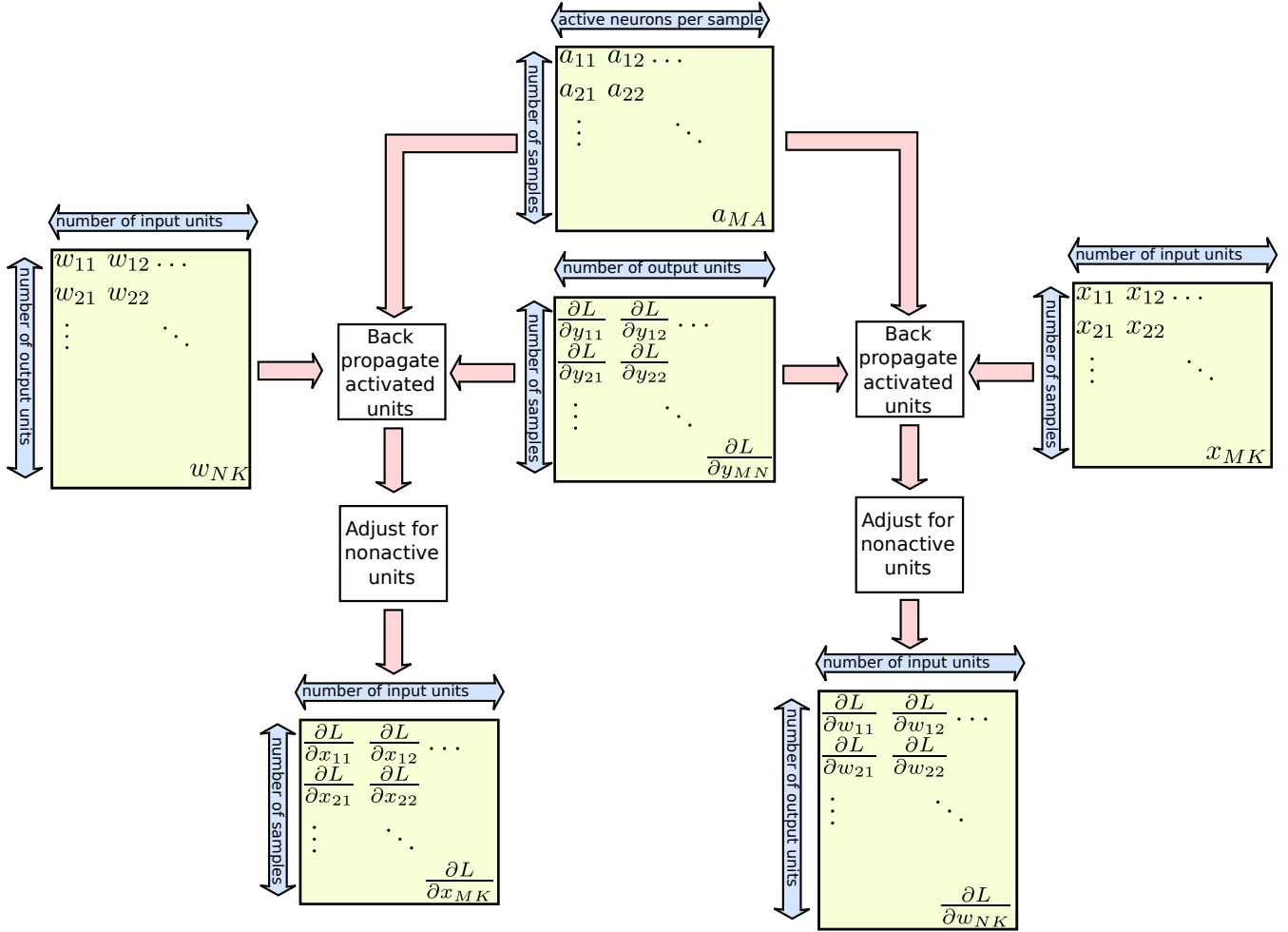}
\caption{The process of finding the choosing the units that lead to the highest results.}
\label{fig_hashedBack}
\end{figure*}

Again computing these two sums of multiplications can be computationally costly given the size of the neural layers. To overcome this we will only use the elements in $\frac{\partial L}{\partial y_{ji}}$ whose corresponding units were activated or have a positive partial derivative as shown in figure~\ref{fig_hashedBack}.

\section{Computational Complexity}
\label{section_complexity}
Each propagation through a fully-connected layer, consists of three multiplications with the sizes $M$,$K$, and $N$, where $M$ is the number of inputs in one batch of samples and $K$ is the size of the input layer and $N$ is the size of the output layer. With this the computational complexity of passing a single batch of samples through a normal neural layer becomes:
\begin{equation}
\label{eq_normalTime}
T_{normal}(M,K,N)=O(MKN).
\end{equation}

In a hashing layer the computational complexity of multiplying the inputs by their selected units is $O(MKA)$ where $A$ is the number of units selected for activation for each sample. In addition to this, for each single sample, we need to compute the hashes and also match the hashes. This has a computational complexity of $O(MQV)$, with $Q$ as the total number of hashes used and $V$ as the average number of votes that a single hash bin assigns. 

In this approach only the computational complexity of placing the output units into the hash table depends on the number of output units. Because we need to hash all units to appropriate bins for all different hashes, the complexity of this operation is $O(NQ)$. With this, the total computation complexity of training on one batch of inputs using the hashed layer becomes:
\begin{equation}
T_{hashed}(M,K,N)=O(MKA+MQV+NQ).
\end{equation}

One significant result of this, compared with the time complexity of a normal layer in equation \ref{eq_normalTime}, is that the variable $N$ has been decoupled from the variables $K$ and $M$. The decoupling from $K$ means that if we increase the size of the input and output units proportionately, the complexity will only increase linearly where as originally it would increase quadratically.

Furthermore the decoupling between $M$ and $N$ means increasing the size of the batches of samples, $M$, in proportion to the number of output units $N$, can result in an amortized value of $Q$ for the term $NQ$ over all the samples processed.

The speed ups obtained in this work are related to both approaches because, in our experiments, the size of the layer next to last is proportional to the size of the last layer. With the explanations given in this section we expect that scaling the size of the last layer even more would lead to the second factor becoming dominant in the amount of savings.

\section{Experiments and results}

The configuration that is commonly used for classification networks is to have a final layer that has the same number of output units as classes. Therefore this layer becomes cumbersome when the number of classes increases and the hashing layer becomes a suitable substitute for this layer.

To validate our proposal, we used a pretrained network to extract the features up to the final layer. The experiment shows that our WTA hashing layer leads to faster training times with competitive accuracy in comparison to the exhaustive matrix multiplication in this last classification layer.

To arrive at these results, we trained the network from~\cite{krizhevsky_imagenet_2012} on the ILSVRC2012 dataset~\cite{russakovsky_imagenet_2014}. We used this trained network to extract features from the imagenet10k~\cite{deng_what_2010}. To do this the test set and the training set were fed through the network and we extracted the FC7 features (4096 features). These features were then used to train a new single output layer of units that would classify the input into the 10184 output categories according to a softmax approach.

The single output layer was trained using two approaches, one was the proposed hashing algorithm and the other using the regular matrix multiplication technique. For the proposed hashing approach we used 8 elements $N_e$ per section and 3 sections, $N_s$, per hash. Furthermore 256 hashes, $Q$, were computed for hashing a given unit.

To perform the test the traditional multiplication technique was used for both networks so that the test approach remains the same. 

\begin{figure}[!t]
\centering
\includegraphics[width=3.25in]{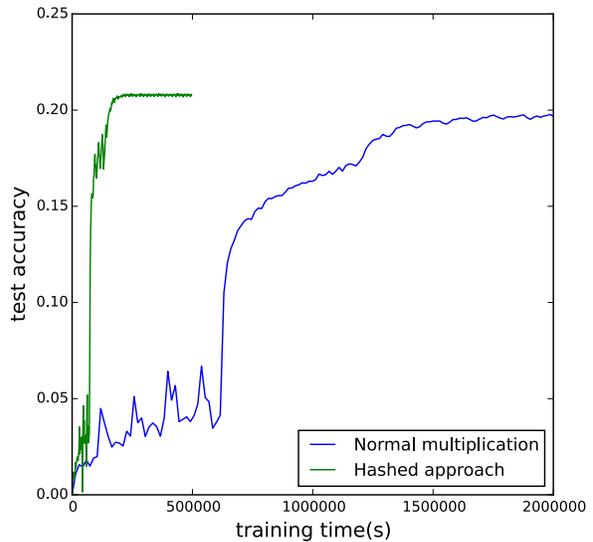}
\caption{Comparing the time required to train a single layer of a classification network on the imagenet10k~\cite{deng_what_2010}. The x axis represents training time and the y axis represent the accuracy of the network.}
\label{fig_singleLayer}
\end{figure}

We tested the network on a separate testing set and recorded the top-1 accuracy ratio. The results, as a function of training time, are shown in figure~\ref{fig_singleLayer}.

From the figure we can see that both approaches behave similarly but our approach converges about 7 times faster. In table \ref{tableTimes} we show a comparison of the computation time of the forward and the backward propagation for the traditional multiplication approach and our proposed hashed technique.

\begin{table}[h]
\begin{tabular}{l|lrlr|}
\cline{2-5}
                                                                                   & \multicolumn{2}{l|}{Normal Neural Layer}                                                                   & \multicolumn{2}{l|}{Hashed Sparse Layer}                                                              \\ \cline{2-5} 
                                                                                   & \multicolumn{1}{l|}{Operation}                                              & \multicolumn{1}{r|}{Time(s)} & \multicolumn{1}{l|}{Operation}                                                              & Time(s) \\ \hline
\multicolumn{1}{|l|}{\multirow{3}{*}{\begin{sideways} activation  \end{sideways}}} & \multicolumn{1}{l|}{\begin{tabular}[c]{@{}l@{}}Weighted\\ Sum\end{tabular}} & \multicolumn{1}{r|}{7.189}   & \multicolumn{1}{l|}{\begin{tabular}[c]{@{}l@{}}Hashed\\ input lookup\end{tabular}}          & 0.378   \\ \cline{2-2} \cline{4-4}
\multicolumn{1}{|l|}{}                                                             & \multicolumn{1}{l|}{}                                                       & \multicolumn{1}{l|}{}        & \multicolumn{1}{l|}{\begin{tabular}[c]{@{}l@{}}Partial\\ Weighted Sum\end{tabular}}         & 1.028   \\ \cline{2-5} 
\multicolumn{1}{|l|}{}                                                             & Total                                                                       & \multicolumn{1}{r|}{7.189}   &                                                                                             & 1.407   \\ \hline
\multicolumn{1}{|l|}{\multirow{3}{*}{\begin{sideways} backprop \end{sideways}}}    & \multicolumn{1}{l|}{\begin{tabular}[c]{@{}l@{}}Weighted\\ Sum\end{tabular}} & \multicolumn{1}{r|}{17.977}  & \multicolumn{1}{l|}{\begin{tabular}[c]{@{}l@{}}Hash weights\\ into hash table\end{tabular}} & 0.248   \\ \cline{2-2} \cline{4-4}
\multicolumn{1}{|l|}{}                                                             & \multicolumn{1}{l|}{}                                                       & \multicolumn{1}{r|}{}        & \multicolumn{1}{l|}{\begin{tabular}[c]{@{}l@{}}Partial \\ Weighted Sum\end{tabular}}        & 1.944   \\ \cline{2-5} 
\multicolumn{1}{|l|}{}                                                             & Total                                                                       & 17.977                       &                                                                                             & 2.243   \\ \hline
                                                                                   & Grand Total                                                                 & \multicolumn{1}{l}{25.166}   &                                                                                             & 3.650   \\ \cline{2-5}

\end{tabular}
\caption{Time required to perform one weight update for one layer of units. Our approach is compared to the normal multiplication approach. The same parameters as the first experiment in the paper is used.}
\label{tableTimes}
\end{table}

To assess the behavior of the hashed layer during forward propagation, one final experiment was performed. We first extracted the FC7 features from the ILSVRC2012 test-set~\cite{russakovsky_imagenet_2014} using the same network as the first experiment. We then copied the weights of the output layer (FC8) from the already trained network to a network that only consists of one hashing layer followed by a softmax classifier.

\begin{figure}[!t]
\centering
\includegraphics[width=3.25in]{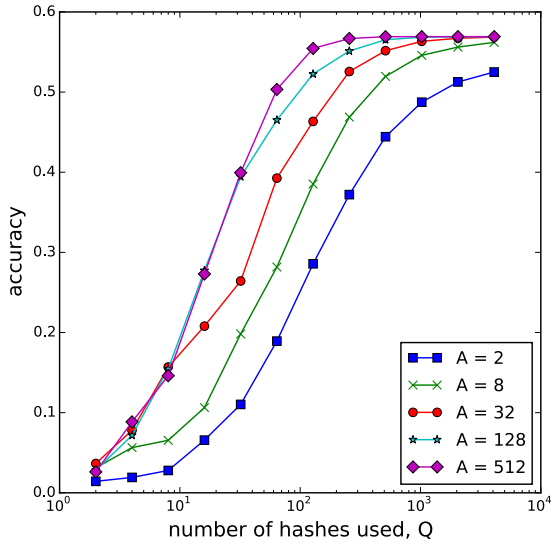}
\caption{The accuracy of the hashed layer on the ILSVRC2012 dataset~\cite{russakovsky_imagenet_2014}. Compared with respect to the number of activated units, $A$, and number of hashes used, $Q$. }
\label{fig_accuracy}
\end{figure}

\begin{figure}[!t]
\centering
\includegraphics[width=3.25in]{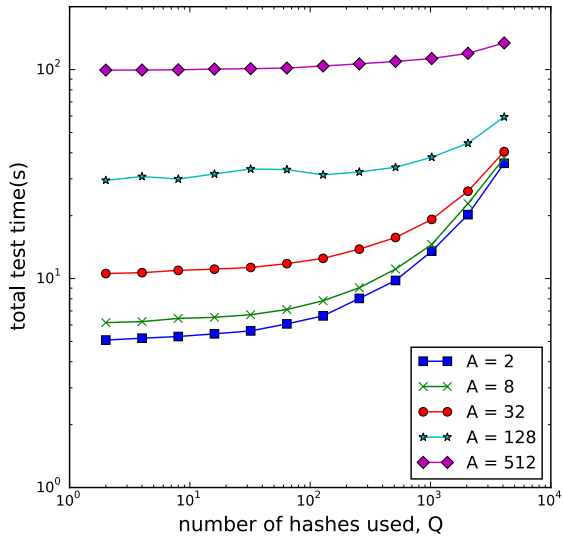}
\caption{The forward propagation time required for the hashing layer on the ILSVRC2012 dataset~\cite{russakovsky_imagenet_2014}, with respect to the number of activated units, $A$, and number of hashes used, $Q$.}
\label{fig_time}
\end{figure}

The new network was used to classify 50000 samples of the ILSVRC2012 testing  dataset using the extracted features. We repeated this several times while changing the main parameters of the hashing layer. The accuracy and computation time of each execution was recorded and they are shown in figure \ref{fig_accuracy} and figure  \ref{fig_time} respectively.

As expected, increasing the number of hashes and active units, both result in increased accuracy at the expense of more required time. But increasing the number of active units results in more penalty because its impact is proportional to the size of the input units. 

On the other hand, the computation of more hashes is relatively cheap and increasing the number of hashes in favor of reducing the number of activated units results in a speedup without compromising performance.

Notice that using 1024 hashes to select 32 active units (which correspond to $3\% $ of the units) results in much less required time while achieving $99.9\%$ of the maximum performance. 

\section{Discussion and conclusions}

In this paper we introduced a hashing approach to speed up the learning of the parameters of the fully-connected layers of a Neural Network. The proposal reduces significantly the computational complexity of both forward and backward propagation. Also, a particular CPU implementation of the algorithm that speeds up the computations by a factor of 7 is introduced.

We will provide access to the CPU implementation used in this paper. We should note that the provided package is not a straight forward implementation of the algorithm given in this article. To compete with the modern matrix multiplication packages and realize the potential of the modern computation machines we needed to attend to memory coalescing, the use of special multi-data, and cache management instructions.

Our ultimate goal is applying the technique to the whole network. Therefore, an important future work is the extension of this approach to conventional layers in CNNs. With this we anticipate achieving the same type of speedup while training the whole network.

In line with our main goal, we are interested in testing the effect of the proposed approach on the all the layers that are fully connected during training. That is we would like to speedup layers 6 and 7, as well as the classification layer.%

Furthermore, our immediate endeavor also includes the completion of our GPU-CUDA implementation. This would allow us to train larger networks, much faster than the current state of the art.

\section{Acknowledgements}
Research supported by TIN2012-38187-C03-02 grant from the Ministerio de Educacion Cultura y Deporte (Spain) and NVidia Hardware grant program.

{\small
\bibliographystyle{ieee}
\bibliography{main}
}

\end{document}